\documentclass{article} 
\usepackage{iclr2027_conference,times}
\usepackage{booktabs}
\usepackage{graphicx}
\usepackage{makecell}

\usepackage{amsmath,amsfonts,bm}

\def\eqref#1{equation~\ref{#1}}

\def\1{\bm{1}}

\DeclareMathAlphabet{\mathsfit}{\encodingdefault}{\sfdefault}{m}{sl}
\SetMathAlphabet{\mathsfit}{bold}{\encodingdefault}{\sfdefault}{bx}{n}

\usepackage{hyperref}
\usepackage{url}
\usepackage{microtype}
\usepackage{multirow}
\definecolor{lightblue}{RGB}{55, 140, 255}

\title{WeLike2Party! In-Context Motion Transfer for Multi-Human Image Animation}

\iclrfinalcopy
\author{
Sangeyl Lee$^{1}$,
Seunghyun Shin$^{2}$,
Seungho Park$^{1}$,
Wooseok Jeon$^{1}$,
Hae-Gon Jeon$^{1}$\thanks{Corresponding author} \\
$^{1}$Department of Artificial Intelligence, Yonsei University \\
$^{2}$AI Graduate School, GIST\\
}

\begin{document}

\maketitle
\vspace{-6mm}

\begin{figure}[h]
  \centering
  \includegraphics[width=\linewidth]{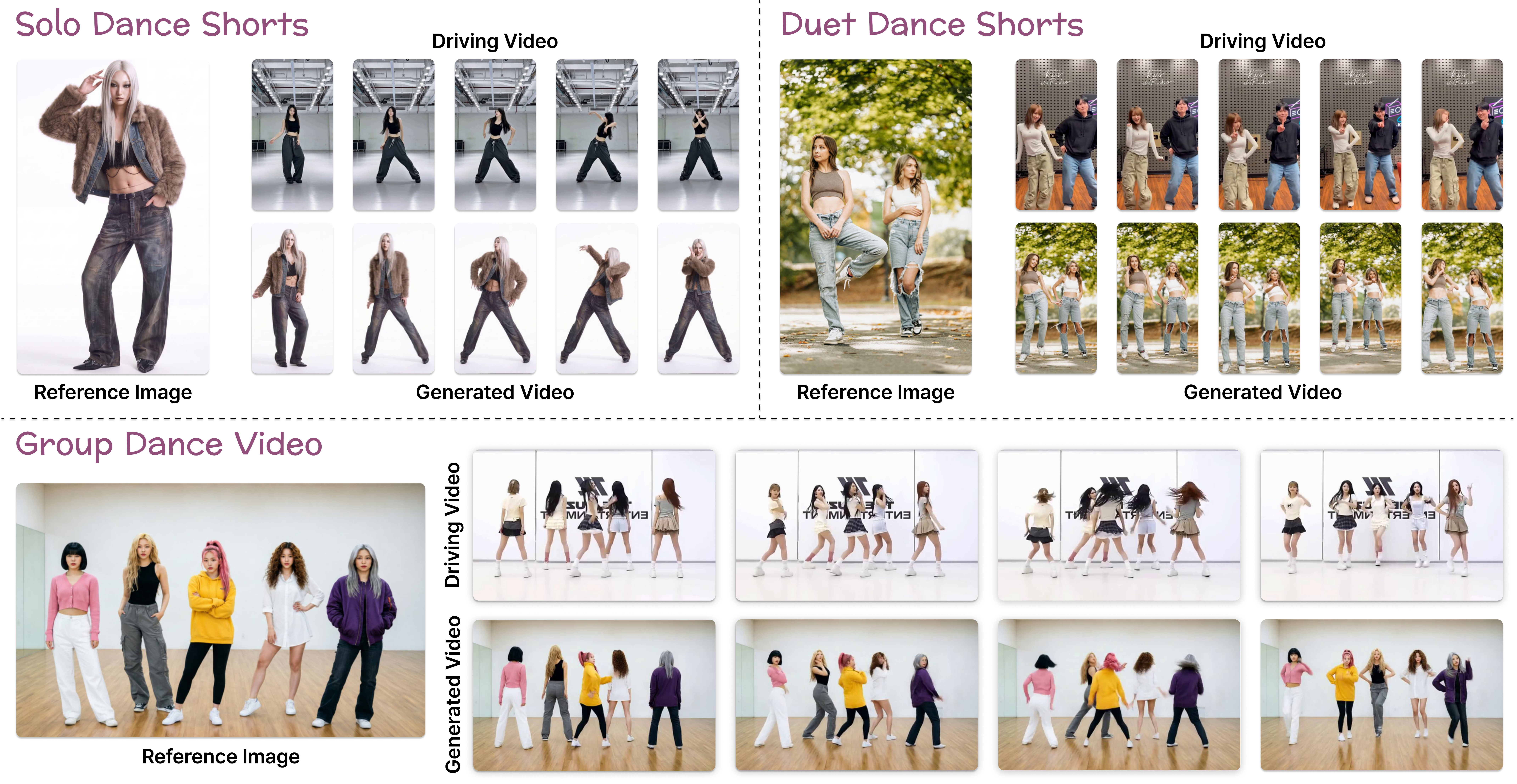}
  \vspace{-4mm}
  \caption{\textbf{Human image animation with WeLike2Party}. Given a reference image and a driving video, WeLike2Party animates the reference subjects while preserving their appearance and identity-motion correspondence. Our framework enables the creation of both solo and group dance videos.}
  \vspace{-1.5mm}
  \label{fig:teaser}
  
\end{figure}

\begin{abstract}

Human image animation aims to transfer motion from a driving video to subjects in a reference image.
Despite remarkable progress in video generation, achieving high-fidelity animation of multiple interacting subjects remains a challenge.
Many existing approaches rely on explicit motion representations such as 2D skeletons or parametric body meshes and struggle to preserve identity-motion binding under inter-person occlusion. 
To address this limitation, we propose \textit{WeLike2Party}, a multi-human animation framework built on direct in-context video conditioning without explicit pose or mesh extraction at inference. 
We further introduce \textit{Reference Asymmetric RoPE Conditioning} to preserve fine-grained appearance details, and \textit{Identity Binding Supervision} to associate each reference identity with its intended motion trajectory.
To support cross-identity training, we construct \textit{MotionTwin}, a large-scale synthetic dataset comprising $14.4$K cross-identity video pairs with shared subject and camera motions, totaling $84.3$ hours of photorealistic video.
We additionally present \textit{MotionTwin-Bench}, a cross-identity benchmark specifically designed to evaluate subject-level visual fidelity and identity-motion binding.  
Extensive experiments on MotionTwin-Bench and real-world videos demonstrate that WeLike2Party outperforms recent state-of-the-art methods in subject-level visual fidelity, identity-motion binding, and overall perceptual quality, particularly in multi-person interactions with substantial occlusion. To validate its robustness, we provide examples of MotionTwin and results on \textbf{\textcolor{lightblue}{\href{https://wl2p-video.github.io/}{Project page}}}.
\end{abstract}

\section{Introduction}

Imagine generating a realistic video of yourself dancing with your favorite performers (See Fig.~\ref{fig:teaser}).
Driven by recent advances in video generation models~\citep{blattmann2023stable, hacohen2024ltx, wan2025}, human image animation enables such scenarios by transferring motion from a driving video to subjects in a reference image while preserving their identity.

Existing approaches~\citep{xu2024magicanimate, hu2024animate, zhang2024mimicmotion, zhu2024champ, wang2025unianimate, tu2025stableanimator} condition video generation on explicit motion representations, such as 2D skeletal maps or parametric 3D human meshes, extracted using off-the-shelf estimators.
These representations allow identity references and motion conditions to be extracted from the same video, enabling self-supervised training without paired cross-identity data.

While these methods achieve promising results, errors in the estimated motion representations can propagate to the generated video and introduce visual artifacts.
These limitations become particularly pronounced in multi-person scenarios, where inter-person occlusion complicates motion estimation and identity association.
For example, skeletal pose estimators can produce missing or incorrectly assigned joints when subjects overlap.
The limited subject-specific information in skeletons further hinders the association between motion cues and individual identities.
Parametric body meshes provide richer geometric information, but frame-level estimation errors can still introduce temporally inconsistent motion guidance.
Moreover, inter-person occlusion can disrupt subject association across frames, causing identity switches in mesh tracking and incorrect identity-motion binding. 

To address these limitations, we propose WeLike2Party~(WL2P), an in-context motion conditioning framework for multi-human image animation that directly transfers motion from a driving video without relying on explicit intermediate representations.
Specifically, we concatenate the visual representations of the reference image, the driving video, and the noisy target video into a single sequence, so that the model can jointly leverage identity, motion, and interaction cues during denoising.
However, naïve in-context conditioning struggles to generate high-quality multi-human videos due to the loss of per-subject appearance detail and the absence of explicit supervision on motion binding. 
 
Therefore, we introduce two components tailored for high-fidelity multi-human motion transfer built on the in-context conditioning framework.
First, Reference Asymmetric RoPE Conditioning~(RARC) incorporates a reference image at a higher resolution than the target video by rescaling its spatial coordinates to the target grid's range before applying Rotary Position Embedding~(RoPE)~\citep{su2024roformer}.
This enables target tokens to retrieve fine-grained details from the reference through self-attention.
Second, Identity Binding Supervision~(IBS) explicitly supervises attention between target and reference image tokens using ground-truth per-subject instance masks, encouraging each reference identity to remain associated with its intended motion trajectory throughout generation.

Training such a direct in-context conditioning model, however, requires paired cross-identity videos that share the same underlying motion while depicting different identities. 
Existing datasets rarely provide this correspondence, particularly for multi-human interactions.
We therefore construct MotionTwin, a large-scale synthetic multi-human dataset containing $84.3$ hours of photorealistic 1080p video organized into $14.4$K cross-identity video pairs. 
Using Unreal Engine 5, each pair is generated by retargeting the same motion onto different combinations of rigged human avatars while independently varying subject identity and scene appearance. 
The dataset draws from $12.4$K single- and multi-person motion sequences, $4.5$K distinct combinations of human assets, and 357 high dynamic range image (HDRI) environments. 
We further apply an automatic filtering pipeline to remove physically implausible interactions and improve the quality of the training pairs.
We additionally introduce MotionTwin-Bench, consisting of 300 cross-identity video pairs from the same pipeline, for evaluating subject-level visual fidelity and identity-motion binding.

We demonstrate the effectiveness of WL2P through comparisons on our benchmark and additional evaluation on diverse real-world image-video pairs using VBench, VBench++\citep{huang2024vbench, huang2025vbench++} and user study.
The results show improvements over recent state-of-the-art methods in per-subject fidelity, motion transfer, identity-motion binding, and overall perceptual quality. 

Our main contributions are as follows:
\vspace{-2.5mm}
\begin{itemize}
    \item We construct MotionTwin, a large-scale synthetic dataset comprising 14.4K cross-identity video pairs spanning diverse single- and multi-person motions. We also establish MotionTwin-Bench for evaluating subject-level visual fidelity and identity-motion binding.
    \item We propose WL2P, an in-context motion transfer framework for multi-human image animation that directly conditions on driving videos. RARC preserves fine-grained appearance details through high-resolution reference encoding and spatial RoPE rescaling, while IBS strengthens identity-motion binding through attention supervision during training. 
    \item Extensive experiments on MotionTwin-Bench and diverse real-world inputs demonstrate that WL2P outperforms recent state-of-the-art methods in subject-level visual fidelity, identity-motion binding, and overall perceptual quality, including challenging multi-person interactions with substantial occlusion.
\end{itemize}

\section{Related Work}
\label{gen_inst}

\paragraph{Controllable video diffusion models.}
Controllable video diffusion models complement text prompts with auxiliary signals that provide fine-grained control over spatial structure, subject motion, appearance, and viewpoint. 
For instance, structural guidance has been introduced through various modalities such as depth, edge, sketch, and mask sequences~\citep{chen2023control, wang2023videocomposer, guo2024sparsectrl}, while human-centric animation methods~\citep{chang2023magicpose, xu2024magicanimate, hu2024animate, zhang2024mimicmotion, zhu2024champ} commonly use 2D poses, facial landmarks, or parametric body models as kinematic priors. 
Reference images and identity embeddings further constrain subject appearance~\citep{jiang2024videobooth, he2024id, shin2025video, tu2025stableanimator}, whereas explicit camera-pose sequences and Pl\"ucker-ray embeddings enable trajectory-level viewpoint control~\citep{he2024cameractrl, zheng2024cami2v, bai2025recammaster, shin2026trimotion}.
More recently, in-context visual conditioning methods directly process tokenized reference or driving videos within diffusion transformers, avoiding the information bottleneck imposed by compact explicit representations~\citep{luo2025camclonemaster, yan2026scail2, wang2026wan}. 
However, directly incorporating rich visual context does not by itself resolve subject-wise correspondence when multiple identities and motions coexist. 
In this work, we focus on this challenge in multi-human image animation, where accurate motion transfer requires preserving the association between each subject identity and its corresponding motion throughout generation.

\paragraph{Human image animation.}
Human image animation aims to synthesize a video of a reference subject following a desired motion sequence. 
Most existing methods encode driving motion using compact explicit representations, such as 2D poses, DensePose, landmarks, or parametric body models, while separately conditioning on reference appearance~\citep{xu2024magicanimate, hu2024animate, zhang2024mimicmotion, zhu2024champ, wang2025unianimate, tu2025stableanimator, cheng2025wan, zhang2025steadydancer}. 
Although this formulation enables self-supervised training, it commonly discards fine-grained geometry and interaction cues, and propagates estimator errors into the generated video, especially for the case where multiple subjects overlap or exchange positions. 
Recent multi-character animation methods mitigate these issues through instance-aware conditions and explicit correspondence mechanisms, including flow and depth guidance, tracking-mask guidance, graph-based matching, identifier-aware representations, and 4D supervision~\citep{xue2025towards, wang2025multi, chen2026dancetogether, ling2026everybodydance, Hu_2026_CVPR, hu2026motionweaver, ding2026mtvcraft}. 
However, these approaches still largely rely on explicit motion representations and their associated estimation pipelines. 
To bypass this dependency, recent works directly condition on driving videos, preserving richer motion and interaction cues at inference~\citep{luo2026dreamactor, yan2026scail2, wang2026wan}.
Nevertheless, these methods are trained on paired datasets synthesized by pose-driven animation models, so their motion correspondence inherits errors from the synthesis pipeline. 
Moreover, SCAIL-2 requires inference-time per-subject masks to specify identity-motion correspondence, while DreamActor-M2 and Wan-Animate 2 provide no explicit mechanism for identity binding and degrade substantially in multi-subject scenarios.

In contrast, we construct multi-human cross-identity training pairs, covering up to seven subjects, by directly retargeting the same underlying motion and camera trajectory to different identities.
This provides ground-truth motion correspondence without relying on pose-driven animation models for data synthesis.
Moreover, WL2P uses per-subject instance masks only as training supervision~(Sec.~\ref{sec:ibs}), requiring neither external pose estimation nor subject-wise binding masks during inference.

\begin{figure}[t]
  \centering
  \vspace{-2mm}
  \includegraphics[width=\linewidth]{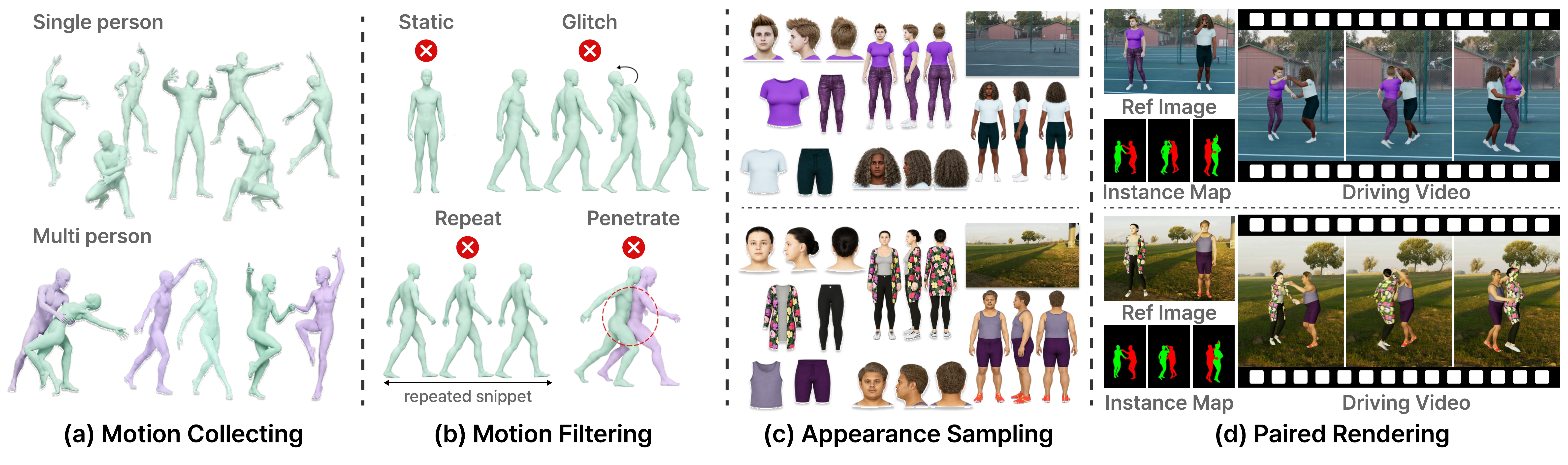}
  \vspace{-9mm}
  \caption{\textbf{MotionTwin curation pipeline.}\quad We collect and filter diverse SMPL-X motions, sample different avatar appearances, and render cross-identity video pairs with shared motion. Multi-person pairs additionally provide reference images and instance maps for identity correspondence.}
  \label{fig:data}
  \vspace{-6mm}
\end{figure}

\section{Dataset: MotionTwin} 
\label{sec:dataset}

Training WL2P requires a cross-identity pair dataset consisting of three components: an identity reference image, a driving video, and a target video. 
Specifically, the driving and target videos share the same pose sequence and camera trajectory while depicting different subjects and scenes, and the identity reference image provides the target subject identities. 
Since collecting such precisely synchronized cross-identity data in the real world is nearly impossible, we construct a large-scale synthetic multi-human dataset using Unreal Engine~5~\citep{epicgames_unrealengine5} as illustrated in Fig.~\ref{fig:data}.

We first collect diverse single and multi-person 3D motion sequences from publicly available human motion datasets~\citep{mahmood2019amass, xu2024inter, li2024duolando, burkanova2025compas3d, fieraru2020three, khirodkar2024harmony4d, yin2023hi4d, le2023music} (See Fig.~\ref{fig:data}(a)).
We then convert them into a unified SMPL-X representation by normalizing the frame rate and coordinate convention. 
Since these datasets originate from heterogeneous capture and reconstruction pipelines, their raw motion quality varies substantially.
Furthermore, they often contain motions unsuitable for animation videos, such as static or excessively traveling movements.

To support realistic and dynamic human animation, we apply an automatic filtering procedure inspired by prior motion curation practices~\citep{fan2025go, lin2023motion} (See Fig.~\ref{fig:data}(b)).
For static segments, we measure the mean per-DoF temporal pose variation and discard the clips when it falls below $10^{-3}$\,radians per frame.
For rotation glitches, we detect them based on frame-to-frame geodesic angular velocity, classifying joint rotations that exceed $60^\circ$ per frame as implausible jumps.
We further constrain each clip to the longest sub-window whose horizontal root-trajectory extent stays within $8$\,m, as larger excursions move beyond a static single-camera field of view.
Lastly, we reduce redundant motion by removing repeated choreography and extracting at most two windows from each long motion sequence.
For multi-person sequences, we additionally filter severe inter-person mesh penetration using distance-field-based collision measures~\citep{jiang2020coherent, ugrinovic2024multiphys}. 
Sequences with maximum penetration depth exceeding $4$ cm are rejected, while shallow contacts from natural interactions are preserved.

After filtering, each motion sequence is rendered into two appearance variants with distinct avatar identities, garments, hairstyles, lighting, and backgrounds (See Fig.~\ref{fig:data}(c)). 
We further introduce body-shape variation between the two variants by symmetrically perturbing the estimated body shape, producing differences in body build, including height and weight. 
This exposes the model to motion correspondence across different body proportions, improving robustness when the driving and target subjects have substantially different body shapes during inference. 
Human appearances are constructed from a diverse asset collection~\citep{black2023bedlam, tesch2026bedlam2}, and scene environments are sampled from $357$ HDRIs~\citep{polyhaven}.

During the rendering procedure, reference images are constructed differently for single- and multi-person sequences (See Fig.~\ref{fig:data}(d)). 
For single-person scenes, the reference image is sampled directly from the target video. 
For multi-person scenes, where occlusion may prevent all subjects from being visible in a single frame, we render reference images by placing subjects side by side following the subject order observed in the first frame of the target video. 
We additionally generate instance masks in Blender~\citep{blender} using the corresponding posed geometry and camera settings to provide subject-level correspondence in Sec.~\ref{sec:ibs}.

In total, MotionTwin contains $14.4$K cross-identity pairs, corresponding to $28.9$K photorealistic videos and $84.3$ hours of content at $1920\times1080$ resolution. 
In particular, our dataset consists of $8.3$K single-person motion pairs and $6.1$K multi-person motion pairs, the latter ranging from two to seven interacting subjects. 
Every multi-person pair is further accompanied by a rest-pose reference image and per-person instance maps, providing $12.3$K reference images with ground-truth identity correspondence between the reference image and the target video. 
These paired videos provide direct supervision for effective motion transfer and identity-motion binding. 

\begin{figure}[t]
  \centering
  \vspace{-4mm}
  \includegraphics[width=\linewidth]{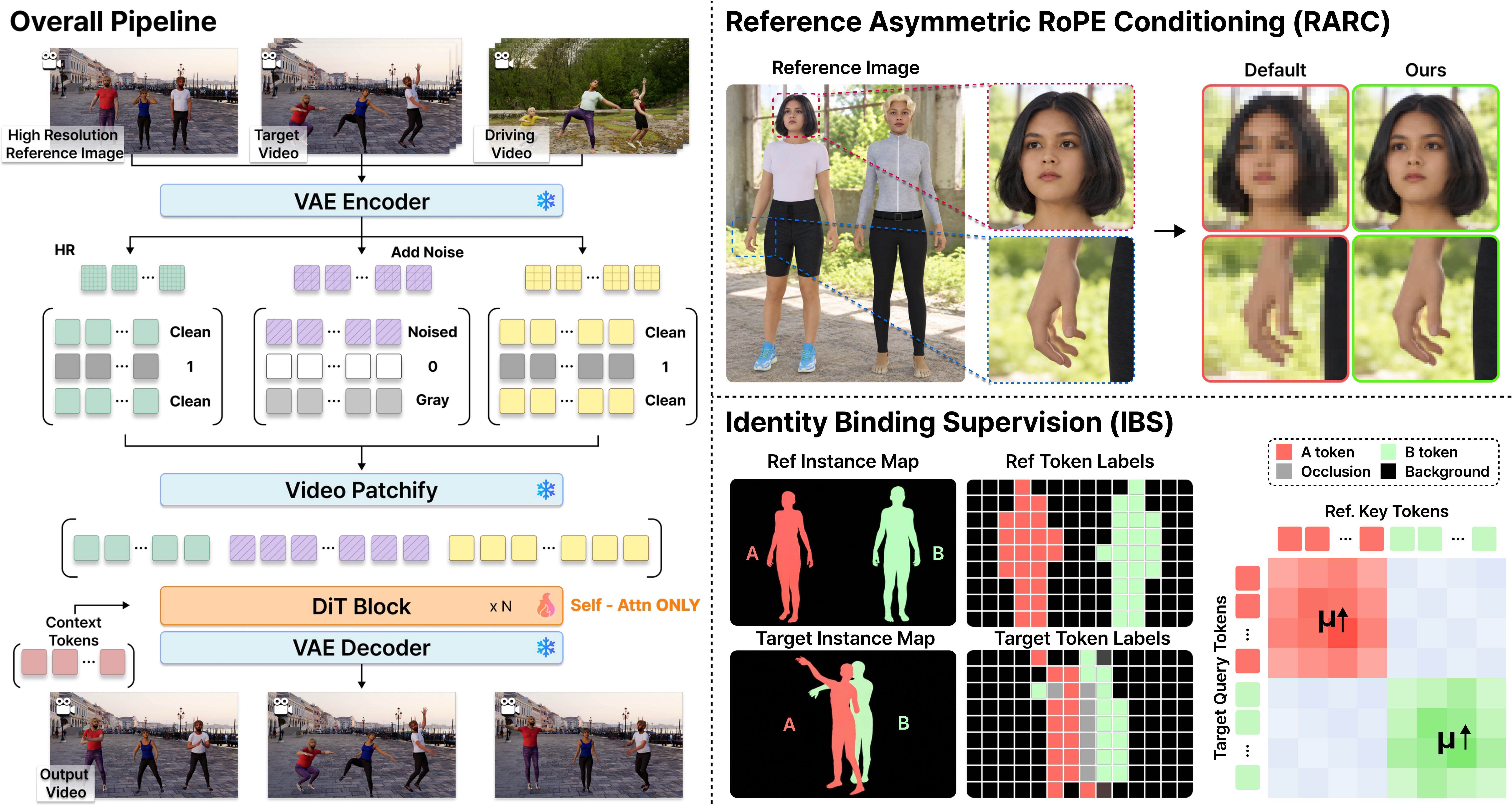}
  \vspace{-5mm}
  \caption{\textbf{Overview of WL2P.}\quad WL2P jointly conditions a DiT on the reference image, driving video, and noisy target video. RARC preserves fine-grained reference details through high-resolution encoding, and IBS enforces subject-level identity-motion correspondence via attention supervision.}
  \vspace{-3mm}
  \label{fig:overview}
\end{figure}

\section{Proposed Method: WL2P}
\label{sec:method}

In this section, we propose WL2P, a framework for multi-human image animation. 
We first introduce the DiT backbone in Sec.~\ref{sec:preliminary} and explain our in-context conditioning layout in Sec.~\ref{sec:in-context}.
We then describe two key components designed for multi-human image animation: (i)~RARC, which improves the representation of fine-grained appearance details~(See Sec.~\ref{sec:rarc}), and (ii)~IBS, which enhances the association between reference identities and their corresponding motion~(See Sec.~\ref{sec:ibs}).
An overview of our framework is illustrated in Fig.~\ref{fig:overview}.

\subsection{Preliminary}
\label{sec:preliminary}

\paragraph{Latent Video Diffusion Transformer.} We build WL2P on Wan2.1-I2V-14B~\citep{wan2025}, an image-conditioned DiT operating in the latent space of a 3D variational autoencoder (VAE)~\citep{kingma2013auto}.
Given an input video $\mathbf{x} \in \mathbb{R}^{3 \times F \times H \times W}$ consisting of $F$ frames, the VAE encoder $\mathcal{E}$ compresses it into a latent $\mathbf{z}_0$ with $C_z=16$ channel dimensions:
\begin{equation}
    \mathbf{z}_0 = \mathcal{E}(\mathbf{x}) \in \mathbb{R}^{C_z \times F' \times H' \times W'},
    \qquad
    F' = \tfrac{F-1}{4} + 1, \quad
    H' = \tfrac{H}{8}, \quad
    W' = \tfrac{W}{8}.
    \label{eq:vae}
\end{equation}
Then, the diffusion forward process adds Gaussian noise $\boldsymbol{\epsilon} \sim \mathcal{N}(\mathbf{0},\mathbf{I})$ to the clean latent $\mathbf{z}_0$ according to the timestep $\tau \in [0,1]$:
\begin{equation}
\mathbf{z}_{\tau}
=
(1-\tau)\mathbf{z}_0+\tau\boldsymbol{\epsilon},
\qquad
\mathbf{u}
=
\boldsymbol{\epsilon}-\mathbf{z}_0,
\end{equation}
where $\mathbf{u}$ denotes a target velocity. 

To incorporate image conditioning, a video with the input image $\mathrm{I}$ in the first frame and zeros in the remaining frames is encoded by $\mathcal{E}$, yielding an image latent $\mathbf{z}_I \in \mathbb{R}^{C_z \times F' \times H' \times W'}$.
To explicitly indicate which latent positions are provided as conditioning, a binary mask is temporally packed into $\mathbf{m} \in \{0,1\}^{4 \times F' \times H' \times W'}$ with ones at the first latent frame and zeros elsewhere.
They are then concatenated channel-wise with the noisy latent $\mathbf{z}_{\tau}$ and projected into a token sequence:
\begin{equation}
    \mathbf{X} = \mathrm{PatchEmbed}([\mathbf{z}_{\tau};\mathbf{m};\mathbf{z}_I]_{\mathrm{ch}}) \in \mathbb{R}^{F'hw\times d},
    \qquad h = \tfrac{H'}{2}, \quad w = \tfrac{W'}{2}
    \label{eq:patchify}
\end{equation}
where $[\cdot]_{\mathrm{ch}}$ denotes channel-wise concatenation and $d$ is the hidden dimension of the DiT.
Given a conditioning context $\mathbf{c}$ for cross-attention, the DiT $\mathbf{v}_{\theta}$ is optimized to predict the target velocity $\mathbf{u}$ using the following flow-matching objective:
\begin{equation}
    \mathcal{L}_{\mathrm{FM}}(\theta)
    =
    \mathbb{E}_{\mathbf{z}_0,\boldsymbol{\epsilon},\tau, \mathbf{c}}
    \left[
        \left\|
            \mathbf{v}_{\theta}
            (\mathbf{X}, \tau, \mathbf{c})
            - \mathbf{u}
        \right\|_2^2
    \right].
    \label{eq:flow_loss}
\end{equation}


\paragraph{3D Rotary Position Embedding.} Our base model employs three-dimensional rotary positional embeddings~(3D-RoPE) to encode the temporal and spatial locations of video tokens within self-attention.
Let a token position be represented as $\mathbf{p}=(p^t,p^h,p^w)$, where $p^t$ indexes the latent frame and $p^h$, $p^w$ denote the height and width coordinates, respectively.
For each coordinate $p$ and RoPE frequency $\omega_k$, a rotation is defined as:
\begin{equation}
    \mathcal{R}(p;\omega_k)
    =
    \begin{bmatrix}
        \cos(p\omega_k) & -\sin(p\omega_k) \\
        \sin(p\omega_k) &  \cos(p\omega_k)
    \end{bmatrix}.
    \label{eq:rope}
\end{equation}
The same rotation is applied to query and key features so that each attention layer can capture relative positional distances between tokens while preserving their content information.

\subsection{In-Context Conditioning}
\label{sec:in-context}

In-context visual conditioning methods inject additional visual context into a DiT by concatenating clean condition tokens with noisy target tokens into a single sequence, allowing self-attention to directly exchange information between them~\citep{luo2025camclonemaster, yan2026scail2, wang2026wan}.
Compared with auxiliary control branches~\citep{zhang2024mimicmotion, wang2025unianimate, cheng2025wan}, this design lets every target token attend directly to every condition token, faithfully transferring identities and motions to the target video.
Thus, we adopt this method to jointly process the identity reference image $\mathrm{I}_{\mathrm{ref}}$, the driving video $\mathrm{V}_{\mathrm{drv}}$, and the target video $\mathrm{V}_{\mathrm{tgt}}$ based on the vanilla Image-to-Video~(I2V) conditioning stream in Eq.~\ref{eq:patchify} without any architectural modification.

Following previous in-context conditioning setups, we treat $\mathrm{I}_{\mathrm{ref}}$ and $\mathrm{V}_{\mathrm{drv}}$ as observed conditions and $\mathrm{V}_{\mathrm{tgt}}$ as the segment to be generated.
Each segment is independently encoded by the VAE encoder $\mathcal{E}$ into a clean latent $\mathbf{z}_{s} \in \mathbb{R}^{C_z \times F'_s \times H'_s \times W'_s}$, where $s \in \{\mathrm{ref}, \mathrm{drv}, \mathrm{tgt}\}$ and $F'_{\mathrm{ref}}=1$.
Here, each segment is set as follows:
\begin{equation}
    (\tilde{\mathbf{z}}_{s},\, \mathbf{m}_{s},\, \tilde{\mathbf{z}}^{I}_{s}) =
    \begin{cases}
        (\mathbf{z}_{s},\ \mathbf{1},\ \mathbf{z}_{s}) & s \in \{\mathrm{ref}, \mathrm{drv}\},\\[2pt]
        (\mathbf{z}_{\mathrm{tgt},\tau},\ \mathbf{0},\ \mathbf{z}_{\emptyset}) & s = \mathrm{tgt},
    \end{cases}
    \label{eq:condmask}
\end{equation}
where $\mathbf{z}_{\mathrm{tgt},\tau}$ is the noisy target latent at timestep $\tau$, $\mathbf{z}_{\emptyset}$ is the VAE encoding of zero-filled frames that the I2V model assigns to target frames, and $\mathbf{1}$ and $\mathbf{0}$ are all-one and all-zero masks of size $4 \times F'_s \times H'_s \times W'_s$.
Since the base model is pretrained with the same masking strategy, $\mathbf{m}_s$ naturally serves as an indicator distinguishing condition tokens from target tokens.

We then patchify each segment and concatenate the resulting tokens along the sequence dimension:
\begin{equation}
    \mathbf{X}_{s} = \mathrm{PatchEmbed}\big([\tilde{\mathbf{z}}_{s};\mathbf{m}_{s};\tilde{\mathbf{z}}^{I}_{s}]_{\mathrm{ch}}\big) \in \mathbb{R}^{N_s \times d},
    \qquad
    \mathbf{X} = [\mathbf{X}_{\mathrm{ref}};\mathbf{X}_{\mathrm{tgt}};\mathbf{X}_{\mathrm{drv}}]_{\mathrm{seq}},
    \label{eq:incontext}
\end{equation}
where $N_s$ is the number of tokens in segment $s$ and $[\cdot]_{\mathrm{seq}}$ denotes sequence-wise concatenation.

\subsection{RARC: Reference Asymmetric RoPE Conditioning}
\label{sec:rarc}

While in-context conditioning effectively transfers the global appearance of the reference image to the target video, fine-grained details such as faces and hands are barely preserved due to the spatial downsampling of the 3D VAE and patch embedding (Eqs.~\ref{eq:vae} and~\ref{eq:patchify}).
A straightforward solution is to encode $\mathrm{I}_{\mathrm{ref}}$ at a higher resolution while keeping the target and driving videos unchanged.
However, directly assigning RoPE coordinates to this denser reference grid places the reference tokens outside the coordinate range observed during pretraining.
Such position extrapolation is known to disrupt the spatial understanding of pretrained DiTs and produce repetitive or implausible content~\citep{zhuo2024lumina, zhao2025riflex}.
We therefore propose RARC, which assigns fractional spatial RoPE coordinates to the reference tokens so that they remain within the spatial range of the target tokens.

Let $h\times w$ denote the spatial token grid of the target and driving videos, and $h_r\times w_r$ the denser grid obtained by encoding $\mathrm{I}_{\mathrm{ref}}$ at a higher resolution.
For a reference token at grid position $(i,j)$, we assign the spatial coordinates as
\begin{equation}
    \tilde{p}^{h}_{i} = i\,\frac{h}{h_r},
    \qquad
    \tilde{p}^{w}_{j} = j\,\frac{w}{w_r},
    \qquad
    i\in\{0,\dots,h_r-1\},\; j\in\{0,\dots,w_r-1\}.
    \label{eq:fractional_rope_pos}
\end{equation}
The pretrained rotation $\mathcal{R}(\tilde{p};\omega_k)$ in Eq.~\ref{eq:rope} is then evaluated directly at these fractional coordinates along the height and width axes.

Under this mapping, segments with different spatial densities share a common RoPE coordinate frame: a target token at $(a,b)$ attends to a reference token at $(i,j)$ with the relative displacement $(a-i\,h/h_r,\ b-j\,w/w_r)$ instead of $(a-i,\ b-j)$.
Consequently, rather than enforcing strict spatial correspondence between the reference image $\mathrm{I}_{\mathrm{ref}}$ and the target video $\mathrm{V}_{\mathrm{tgt}}$, RARC provides spatial cues useful for self-attention to learn which appearance details to transfer to each target region.

\subsection{IBS: Identity Binding Supervision}
\label{sec:ibs}
The flow-matching objective in Eq.~\ref{eq:flow_loss} does not specify which reference identity should correspond to each target subject.
This ambiguity becomes more problematic in multi-human animation, where multiple subjects may overlap or exchange positions with each other.
As a result, the model may produce identity swaps or drifts, where the identity of one reference subject is attached to the motion of another or gradually blends with an interacting subject.
To address this issue, we introduce IBS, which directly supervises reference-to-target attention using ground-truth instance masks from MotionTwin to preserve subject-level identity association.

\paragraph{Subject labels.}
IBS is applied to multi-person pairs, for which MotionTwin provides instance maps of P subjects in both $\mathrm{I}_{\mathrm{ref}}$ and $\mathrm{V}_{\mathrm{tgt}}$.
We first downsample the instance masks to the reference and target token grids.
For each token $x$, let $\ell(x)\in\{0,1,\dots,P\}$ be the subject that occupies the largest area of its region, where $0$ denotes the background and $c(x)\in[0,1]$ the fraction of the region covered by that subject.
We discard background tokens and tokens with $c(x)<\gamma$, which mostly lie on subject boundaries or occluded areas.
The remaining tokens form the set of labeled target tokens $\mathcal{Q}$ and, for each subject $p$, the set of labeled reference tokens $\mathcal{K}_p$, whose union is denoted by $\mathcal{K}$.

\paragraph{Binding loss.}
For target token $u\in\mathcal{Q}$ and reference token $v\in\mathcal{K}$, we compute the attention over the labeled reference tokens and the probability of attending to the correct subject:
\begin{equation}
    a^{l,m}_{uv}
    =
    \frac{\exp\big(\langle \mathbf{q}^{l,m}_{u}, \mathbf{k}^{l,m}_{v} \rangle / \sqrt{d_h}\big)}
    {\sum_{v'\in\mathcal{K}} \exp\big(\langle \mathbf{q}^{l,m}_{u}, \mathbf{k}^{l,m}_{v'} \rangle / \sqrt{d_h}\big)},
    \qquad
    \mu^{l,m}_{u}
    =
    \sum_{v\in\mathcal{K}_{\ell(u)}} a^{l,m}_{uv},
    \label{eq:ibs_attention}
\end{equation}
where $\mathbf{q}$ and $\mathbf{k}$ are the query and key at block $l$ and head $m$, and $d_h$ is the head dimension.
IBS maximizes $\mu^{l,m}_{u}$ weighted by the label confidence:
\begin{equation}
    \mathcal{L}_{\mathrm{IBS}}
    =
    -\frac{1}{|\mathcal{B}|\,M\,|\mathcal{Q}|}
    \sum_{l\in\mathcal{B}}
    \sum_{m=1}^{M}
    \sum_{u\in\mathcal{Q}}
    c(u)\,\log \mu^{l,m}_{u},
    \label{eq:ibs_loss}
\end{equation}
where $\mathcal{B}$ is the set of supervised blocks and $M$ is the number of heads.
By normalizing attention only over reference tokens, IBS controls how attention is distributed among different identities rather than increasing the overall reliance on the reference image.
It does not require pixel-level alignment between the reference image and target video, since only subject-level correspondence is supervised. 
We apply IBS only at high noise levels since coarse subject layouts are mainly determined in early denoising stages~\citep{wu2024freeinit, choi2022perception}.
Note that IBS is used only during training and does not introduce neither additional parameters nor inference cost.
The overall training objective combines the flow-matching loss in Eq.~\ref{eq:flow_loss} with the IBS loss weighted by $\beta$ as: 
\begin{equation}
    \mathcal{L}
    =
    \mathcal{L}_{\mathrm{FM}}
    +
    \beta\, \mathcal{L}_{\mathrm{IBS}}.
    \label{eq:total_loss}
\end{equation}

\vspace{-6mm}
\section{Experimental Results}

\subsection{Implementation Details}
\label{sec:setup}

We train only the self-attention blocks of the Wan2.1-I2V-14B backbone while keeping the remaining parameters frozen.
Training is performed on four NVIDIA B200 GPUs using BF16 mixed precision.
The reference image is resized to a $720$\-pixel short side, while the target and driving videos contain $81$ frames at $640\times352$ resolution for training and $832\times480$ resolution for inference. 
We optimize the model with AdamW for $34$K steps, using a learning rate of $1\times10^{-5}$.
At inference, an $81$-frame $480$p clip takes 6.5 minutes on one B200 GPU.
For IBS, we set $\gamma=0.7$ and $\beta=0.05$ and restrict IBS to high-noise timesteps, corresponding to the top $40\%$ of the noise range.
Additional details are provided in Appendix~\ref{app:training_details}.

\begin{table}[t]
    \centering
    \vspace{-4mm}
    \caption{
        Quantitative evaluation compared to state-of-the-art methods on MotionTwin-Bench.
    }
    \vspace{1mm}
    \label{tab:bindjudge}
    \setlength{\tabcolsep}{4.0pt}
    \renewcommand{\arraystretch}{1.15}
    \resizebox{\linewidth}{!}{%
    \begin{tabular}{l|cccc|ccc|cc}
        \toprule
        \multirow{2}{*}{\vspace{-5pt}Method}
        & \multicolumn{4}{c|}{Full-frame Fidelity}
        & \multicolumn{3}{c|}{Subject Fidelity}
        & \multicolumn{2}{c}{Identity Binding} \\
        \cmidrule{2-10}
        & \makecell{PSNR$\uparrow$}
        & \makecell{SSIM$\uparrow$}
        & \makecell{LPIPS$\downarrow$}
        & \makecell{FVD$\downarrow$}
        & \makecell{mPSNR$\uparrow$}
        & \makecell{mSSIM$\uparrow$}
        & \makecell{mLPIPS$\downarrow$}
        & \makecell{~~IAA$\uparrow$~~}
        & \makecell{~~IAA$_{\text{cross}}$$\uparrow$~~}
        \\
        \midrule

        MultiAnimate~\citep{Hu_2026_CVPR}~~
        & \underline{20.22}
        & \underline{0.7112}
        & \underline{0.3283}
        & 226.61
        & \underline{18.41}
        & \underline{0.7326}
        & \underline{0.2029}
        & 0.8360
        & 0.8192 \\

        Wan-Animate 2~\citep{wang2026wan}~~
        & 17.18
        & 0.5957
        & 0.4609
        & 277.51
        & 15.91
        & 0.6786
        & 0.2969
        & 0.6035
        & 0.5955 \\

        SCAIL~\citep{yan2026scail}~~
        & 17.42
        & 0.5681
        & 0.3856
        & 164.33
        & 18.11
        & 0.7192
        & 0.2188
        & \underline{0.8514}
        & \underline{0.8422} \\

        SCAIL-2~\citep{yan2026scail2}~~
        & 17.68
        & 0.5775
        & 0.3659
        & \underline{120.75}
        & 17.55
        & 0.7097
        & 0.2334
        & 0.8406
        & 0.8335 \\

        \midrule
        \textbf{WeLike2Party~(Ours)}
        & \textbf{25.14}
        & \textbf{0.7939}
        & \textbf{0.1123}
        & \textbf{35.51}
        & \textbf{25.38}
        & \textbf{0.8685}
        & \textbf{0.0691}
        & \textbf{0.9569}
        & \textbf{0.9496} \\

        \bottomrule
    \end{tabular}%
    }
    \vspace{-6mm}
\end{table}

\subsection{Evaluation Protocol}
\label{sec:benchmark}

We evaluate identity-motion binding and visual fidelity on MotionTwin-Bench and real-world inputs.

\noindent\textbf{MotionTwin-Bench.}
To evaluate reconstruction fidelity and identity-motion binding against ground truth, we construct a benchmark of $300$ cross-identity video pairs using the rendering pipeline in Sec.~\ref{sec:dataset}.
We report PSNR, SSIM~\citep{wang2004image}, LPIPS~\citep{zhang2018unreasonable} and FVD~\citep{unterthiner2018towards} for overall video fidelity.
Subject-level fidelity is measured with mPSNR, mSSIM, and mLPIPS using ground-truth per-subject instance masks.
We further introduce Identity Assignment Accuracy ($\mathrm{IAA}$) to assess identity-motion binding.
Using ground-truth instance masks, we match generated subject regions to reference identities based on color-histogram distances, assigning each frame a score of $1$ if all assignments are correct and $0$ otherwise.
$\mathrm{IAA}$ averages these scores over all frames, while the occlusion-focused variant, $\mathrm{IAA}_{\mathrm{cross}}$, considers only frames including inter-person occlusion.
Further details are provided in Appendix~\ref{app:benchmark}.

\noindent\textbf{Real-world evaluation.}
To assess performance beyond the rendered benchmark, we evaluate on real-world inputs.
For quantitative comparison, we use $161$ internet-collected image-video pairs containing both single and multi subjects and report video and frame quality from VBench~\citep{huang2024vbench} and I2V quality from VBench++~\citep{huang2025vbench++}.
These measures assess perceptual quality and consistency with the reference image.
Qualitative comparisons and a user study provide complementary assessments of individual identity preservation and motion assignment.

\noindent\textbf{Compared methods.}
We compare WL2P with recent open-source character animation models, including SCAIL~\citep{yan2026scail}, SCAIL-2~\citep{yan2026scail2}, Wan-Animate 2~\citep{wang2026wan}, and MultiAnimate~\citep{Hu_2026_CVPR}.
Following the official implementation, we generate per-subject masks with SAM3~\citep{carion2026sam} for MultiAnimate and SCAIL-2. 

\subsection{Comparison with State-of-the-Art Methods}

\noindent\textbf{Quantitative evaluations.}
As shown in Tab.~\ref{tab:bindjudge}, WL2P achieves the best results across reconstruction fidelity and identity-motion binding on the benchmark.
MultiAnimate provides the strongest image reconstruction results among the baselines, whereas SCAIL attains the highest binding accuracy.
WL2P improves both aspects jointly, with the gains in masked metrics confirming better reconstruction within subject regions.
Its lower FVD indicates closer agreement with the target video distribution, and its advantage in $\mathrm{IAA}_{\text{cross}}$ supports more reliable identity assignment under inter-person occlusion.

The real-world results in Tab.~\ref{tab:xdance_vbench} show that WL2P also leads in video, frame, and I2V quality.
The improvement in reference consistency along with perceptual quality supports the robustness of our model, which is also applicable to real inputs.

\begin{table}[t]
    \centering
    \vspace{-1mm}
    \caption{
        Quantitative evaluation compared to state-of-the-art methods on the real-world scenario. We report mean and standard deviation for the user study.
    }
    \label{tab:xdance_vbench}
    \setlength{\tabcolsep}{4.0pt}
    \renewcommand{\arraystretch}{1.15}
    \resizebox{\linewidth}{!}{%
    \begin{tabular}{l|cc|c|ccc}
        \toprule
        \multirow{2}{*}{\vspace{-5pt}Method}
        & \multicolumn{2}{c|}{VBench}
        & VBench++
        & \multicolumn{3}{c}{User Study} \\
        \cmidrule{2-7}
        & \makecell{V-Quality$\uparrow$}
        & \makecell{F-Quality$\uparrow$}
        & \makecell{I2V-Quality$\uparrow$}
        & \makecell{Id-Motion Bind$\uparrow$} 
        & \makecell{Id Preserve$\uparrow$} 
        & \makecell{Visual Quality$\uparrow$}
        \\
        \midrule
        MultiAnimate~\citep{Hu_2026_CVPR}
        & 79.99
        & 81.40
        & 85.62
        & 2.575 $\pm$ 1.324
        & 2.355 $\pm$ 1.435
        & 2.480 $\pm$ 1.317\\

        Wan-Animate 2~\citep{wang2026wan}~~
        & 81.58
        & 83.09
        & 87.61
        & 2.385 $\pm$ 1.188
        & 2.558 $\pm$ 1.086
        & 2.430 $\pm$ 1.174 \\

        SCAIL~\citep{yan2026scail}
        & \underline{82.84}
        & 83.61
        & 85.91
        & \underline{3.153} $\pm$ 1.473
        & \underline{3.338} $\pm$ 1.408
        & \underline{3.205} $\pm$ 1.431 \\

        SCAIL-2~\citep{yan2026scail2}
        & 82.70
        & \underline{84.13}
        & \underline{88.41}
        & 2.845 $\pm$ 1.212
        & 2.728 $\pm$ 1.224
        & 2.860 $\pm$ 1.253\\

        \midrule
        \textbf{WeLike2Party~(Ours)}
        & \textbf{83.26}
        & \textbf{84.63}
        & \textbf{88.73}
        & \textbf{4.043} $\pm$ 1.235
        & \textbf{4.022} $\pm$ 1.206
        & \textbf{4.025} $\pm$ 1.258\\
        \bottomrule
    \end{tabular}%
    }
    \vspace{-3mm}
\end{table}

\begin{figure}[t]
  \centering
  \includegraphics[width=0.96\linewidth]{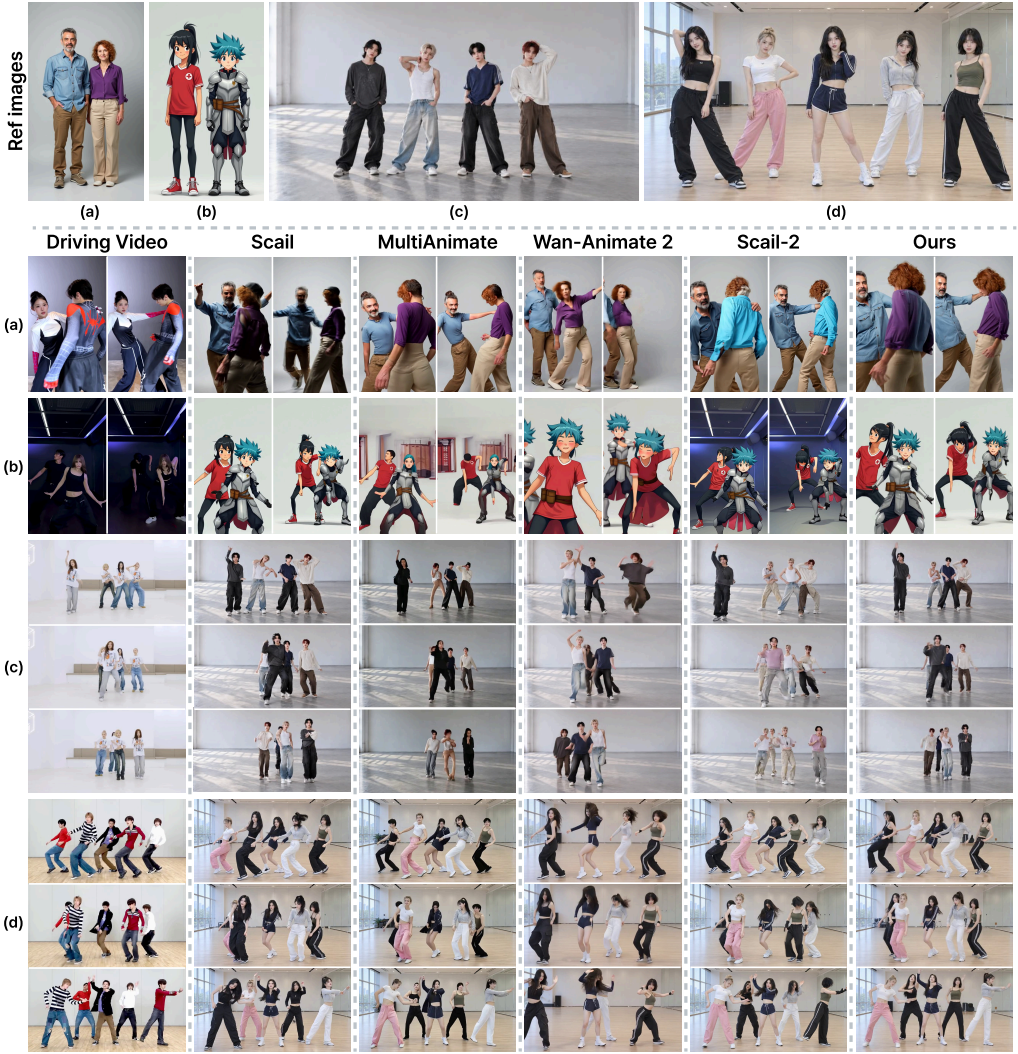}
  \vspace{-4mm}
  \caption{\textbf{Qualitative comparison.} MultiAnimate transfers the driving subjects' hairstyles and faces (a to d), Wan-Animate 2 merges or drops subjects (b to d), SCAIL swaps identities (c, d), and SCAIL-2 copies the driving background (b), alters clothing colors (a, c), and duplicates subjects (c). WL2P preserves each reference subject's appearance and binds it to the intended motion.}
  \vspace{-4mm}
  \label{fig:qualit1}
\end{figure}

\noindent\textbf{Qualitative evaluations.}
Fig.~\ref{fig:qualit1} presents qualitative comparisons for single- and multi-person animation, focusing on subject-level visual fidelity and identity-motion binding.
MultiAnimate tends to reproduce the body proportions of the driving subjects rather than those of the reference subjects and exhibits noticeable appearance degradation. 
Wan-Animate 2 frequently suffers from identity bleeding, with distinct subjects blending into a single figure. 
SCAIL and SCAIL-2 produce the most visually faithful results among the baselines in these examples, but remain susceptible to identity drift as the number of subjects increases and inter-person occlusions become more complex.
In contrast, WL2P shows high-fidelity multi-human animations, preserving each reference subject's appearance and maintaining its association with the intended motion throughout these challenging interactions.

\noindent\textbf{User study.}
We conduct a user study with $20$ participants through Amazon Mechanical Turk (MTurk)~\citep{crowston2012amazon} to evaluate $20$ real-world video clips.
For each clip, participants view the reference image, driving video, and outputs from WL2P and the four baselines.
The outputs are anonymized and displayed simultaneously in randomized order.
Participants rank the five methods from $1$ (best) to $5$ (worst) for identity preservation, identity-motion binding, and overall visual quality.
We convert ranks to scores from $5$ (best) to $1$ (worst) and average them across clips and participants.
As shown in Tab.~\ref{tab:xdance_vbench}, WL2P achieves the highest mean preference scores across all three criteria. 
These results support that WL2P produces user-favored animations while preserving each reference identity and its association with the intended motion.

\subsection{Ablation Studies}
\begin{table}[t]
    \centering
    \footnotesize
    \vspace{-4mm}
    
    \caption{
        Ablation study on RARC and IBS. ``Base + HR ref.'' uses high-resolution reference images with standard RoPE and no coordinate rescaling. 
    }
    \vspace{1mm}
    \label{tab:abl}
    \setlength{\tabcolsep}{4.0pt}
    \renewcommand{\arraystretch}{1.15}
    \resizebox{\linewidth}{!}{%
    \begin{tabular}{l|cccc|ccc|cc}
        \toprule
        \multirow{2}{*}{\vspace{-5pt}Method}
        & \multicolumn{4}{c|}{Full-frame Fidelity}
        & \multicolumn{3}{c|}{Subject Fidelity}
        & \multicolumn{2}{c}{Identity Binding} \\
        \cmidrule{2-10}
        & \makecell{~~~PSNR$\uparrow$~~~}
        & \makecell{~~~SSIM$\uparrow$~~~}
        & \makecell{~~~LPIPS$\downarrow$}
        & \makecell{~~~FVD$\downarrow$~~~}
        & \makecell{~~~mPSNR$\uparrow$~~~}
        & \makecell{~~~mSSIM$\uparrow$~~~}
        & \makecell{~~~mLPIPS$\downarrow$~~~}
        & \makecell{~~~IAA$\uparrow$~~~}
        & \makecell{~~~IAA$_{\text{cross}}$$\uparrow$~~~}
        \\
        \midrule
        Base
        & 22.47
        & 0.7022
        & 0.1542
        & 48.81
        & 22.92
        & 0.8122
        & 0.1232
        & 0.8982
        & 0.8891 \\

        Base + RARC
        & 23.61
        & 0.7090
        & 0.1264
        & 44.07
        & 24.21
        & 0.8487
        & 0.0851
        & 0.9241
        & 0.9171 \\
        Base + HR ref.
        & 22.50
        & 0.6655
        & 0.1395
        & 46.21
        & 23.55
        & 0.8396
        & 0.0947
        & 0.8969
        & 0.8900 \\
        Base + IBS
        & \underline{24.49}
        & \underline{0.7419}
        & \underline{0.1156}
        & \underline{36.55}
        & \underline{24.70}
        & \underline{0.8581}
        & \underline{0.0772}
        & \underline{0.9303}
        & \underline{0.9210} \\
        \midrule
        \textbf{Full (Ours)}
        & \textbf{25.14}
        & \textbf{0.7939}
        & \textbf{0.1123}
        & \textbf{35.51}
        & \textbf{25.38}
        & \textbf{0.8685}
        & \textbf{0.0691}
        & \textbf{0.9569}
        & \textbf{0.9496} \\

        \bottomrule
    \end{tabular}%
    }
    \vspace{-6mm}
\end{table}

To assess the impact of RARC and IBS, we conduct ablation studies on MotionTwin-Bench as shown in Tab.~\ref{tab:abl}.
First, we train a vanilla in-context model as the baseline and evaluate the effect of adding each component separately.
Notably, this baseline already achieves competitive subject fidelity and identity-motion binding, suggesting that useful subject correspondences can emerge from paired training alone.
Next, we compare two variants trained with the same high-resolution reference images: RARC and standard RoPE without coordinate rescaling.
RARC improves subject fidelity by exploiting finer reference details, whereas standard RoPE performs comparably to or worse than the baseline.
We attribute this gap to positional distribution shifts caused by reference resolution changes between training and inference.
RARC mitigates this mismatch through fractional coordinate rescaling, as discussed in Appendix~\ref{sec:additional_experiments}.
Finally, IBS improves both $\mathrm{IAA}$ and $\mathrm{IAA}_{\mathrm{cross}}$ over the baseline, showing that explicit attention supervision further strengthens identity-motion binding.

\section{Conclusion}

We present WeLike2Party, an in-context framework for multi-human image animation that transfers motion directly from driving videos without explicit pose or mesh estimation.
Within this framework, RARC combines high-resolution reference encoding with fractional spatial RoPE to preserve fine-grained appearance details. 
IBS introduces an auxiliary loss on target-to-reference attention to encourage correct identity-motion binding, using per-subject instance masks. 
Furthermore, we introduce MotionTwin to support paired cross-identity training and MotionTwin-Bench to evaluate subject-level visual fidelity and identity-motion binding with ground-truth targets and correspondences. 
Together, these contributions bring the idea of dancing with your favorite performers closer to reality, with each subject retaining their distinct appearance and following the intended motion.

\noindent\textbf{Limitations and Future Works.}
While WeLike2Party produces high-quality multi-human animations across diverse subjects and motions, there is still room for improvement.
First, fine-grained identity details may degrade as more subjects share a fixed-resolution output frame, leaving fewer pixels per person.
Subject-specific face embeddings could help preserve facial identity under dynamic motion and at small on-screen scales.
Second, WL2P remains computationally expensive, which limits its applicability to interactive animation.
Future work could explore efficient attention mechanisms and model distillation to reduce inference cost while preserving appearance fidelity and identity-motion binding.
Another promising direction is to extend MotionTwin with independently controlled camera trajectories and subject motions, providing structured supervision for models that support separate control over camera and human motion.

\clearpage

\subsection*{AI use statement}

In this work, we used generative AI tools to polish the manuscript for readability and grammar.
We have not used generative AI tools to develop the research methodology, implement the proposed method, or interpret experimental results. 
All AI-assisted edits were reviewed and revised by the authors to ensure technical correctness and consistency with the intended claims. 
We take responsibility for the final content of this work, including text, claims or artifacts produced with the aid of generative AI.

\subsection*{Ethics statement}

Our work focuses on multi-human image animation and introduces a synthetic dataset constructed from existing motion resources and rendered human avatars.
The dataset construction process does not involve collecting new personally identifiable information from human subjects.
For real-world evaluation and qualitative figures, we use driving videos collected from publicly available online sources and reference images that are either synthetically generated or drawn from public sources. These materials are used solely for research evaluation and illustration, are not part of the released resources, and will be removed upon request from the individuals depicted or the rights holders.
The proposed method and released resources are intended for research purposes only.
As with other human image and video generation methods, the model could potentially be misused to create misleading or unauthorized content.
We therefore encourage responsible use and careful consideration of consent, privacy, and applicable licenses when using reference images and driving videos.

\subsection*{Reproducibility statement}

For reproducibility, we provide detailed descriptions of the WL2P architecture, training objectives, and MotionTwin construction pipeline in the paper.
The training and evaluation code, model checkpoints, and benchmark data will be released after the final decision.

\clearpage

\bibliography{iclr2027_conference}
\bibliographystyle{iclr2027_conference}

\clearpage
\appendix
\section{Additional Details on MotionTwin Dataset}

\begin{figure}[p] 
    \centering
    \includegraphics[height=0.93\textheight,keepaspectratio]{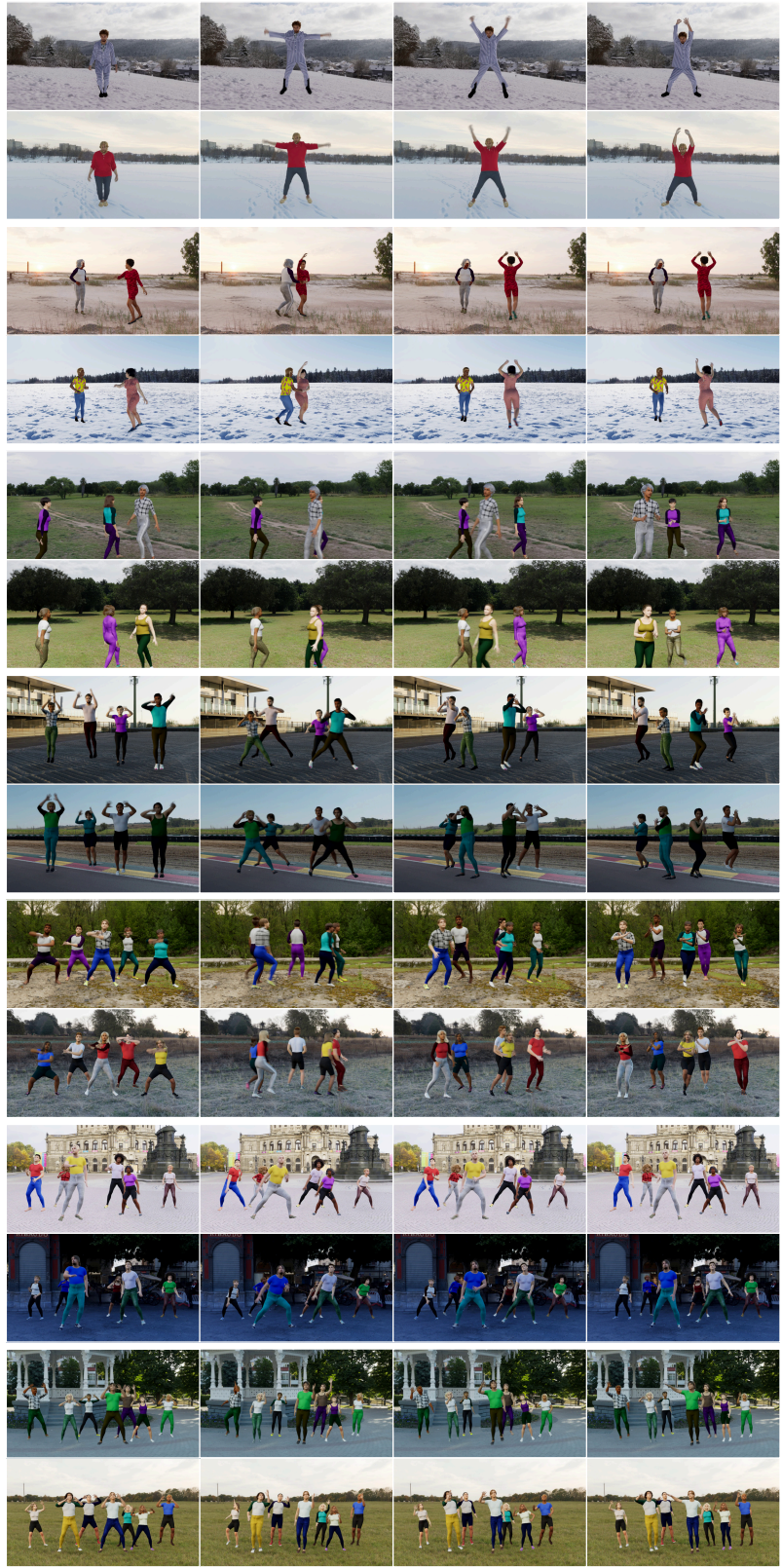}
    \vspace{-1mm}
    \caption{\textbf{Overview of MotionTwin.} Samples with 1–7 people across diverse scenes.}
    \label{fig:motiontwin_overview}
\end{figure}

In this section, we provide additional details on our MotionTwin dataset. An overview of MotionTwin is depicted in Fig.~\ref{fig:motiontwin_overview}.

\begin{figure}[t]
    \centering
    \begin{minipage}[c]{0.40\linewidth}
    \centering
    \includegraphics[width=\linewidth]{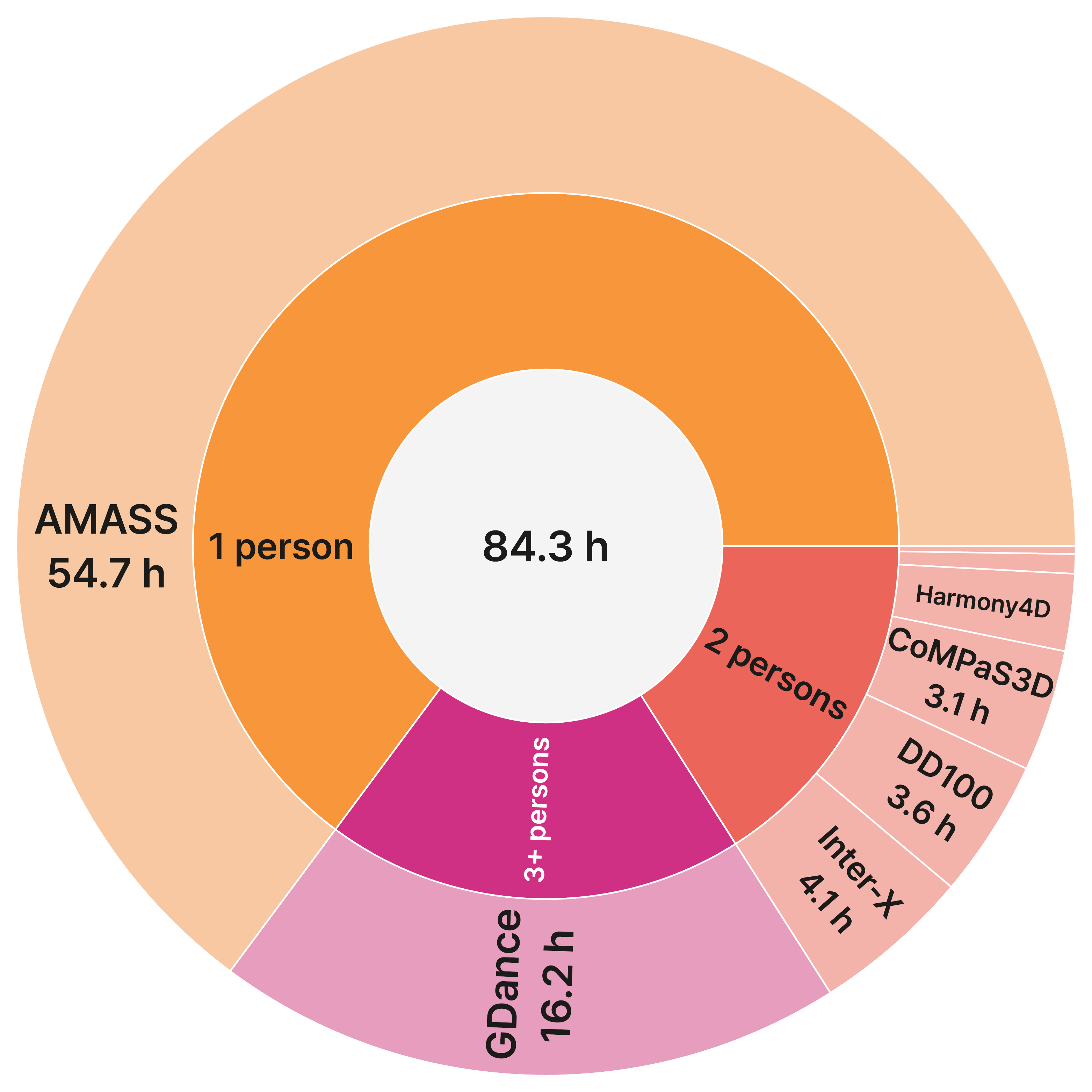}
    \end{minipage}\hfill
    \begin{minipage}[c]{0.57\linewidth}
    \centering
    \scriptsize
    \setlength{\tabcolsep}{7pt}
    \renewcommand{\arraystretch}{1.15}
    \begin{tabular}{lcrrr}
    \toprule
    \textbf{Source}
    & \textbf{Subjects}
    & \textbf{Clips}
    & \textbf{Hours}
    & \textbf{Mean len.\ (s)} \\
    \midrule
    AMASS      & 1    & 8,270  & 54.7 & 11.9 \\
    Inter-X    & 2    & 1,189  & 4.1  & 6.2  \\
    DD100      & 2    & 780    & 3.6  & 8.2  \\
    CoMPaS3D   & 2    & 716    & 3.1  & 7.9  \\
    Harmony4D  & 2    & 248    & 2.0  & 14.4 \\
    CHI3D      & 2    & 253    & 0.5  & 3.6  \\
    Hi4D       & 2    & 94     & 0.2  & 4.0  \\
    GDance     & 2--7 & 2,877  & 16.2 & 10.1 \\
    \midrule
    \textbf{MotionTwin} & 1--7 & \textbf{14,427} & \textbf{84.3} & 10.5 \\
    \bottomrule
\end{tabular}
    \end{minipage}
    \vspace{-3mm}
    \caption{\textbf{MotionTwin statistics.} Left: hours of rendered video by group size (inner ring) and motion source (outer ring). Right: clips and hours per source.}
    \label{fig:data_dist}
\end{figure}

\paragraph{Data distribution.}
MotionTwin contains $14{,}427$ motion clips, each rendered twice with different identities, for a total of $84.3$\,h of rendered video. 
As shown in Fig.~\ref{fig:data_dist}, the motions are drawn from eight public human motion corpora:
single-person clips from AMASS~\citep{mahmood2019amass} ($8{,}270$ clips, $54.7$\,h), two-person clips from Inter-X~\citep{xu2024inter}, DD100~\citep{li2024duolando}, CoMPaS3D~\citep{burkanova2025compas3d}, CHI3D~\citep{fieraru2020three}, Harmony4D~\citep{khirodkar2024harmony4d}, and Hi4D~\citep{yin2023hi4d} ($3{,}280$ clips, $13.5$\,h in total), and clips with three to seven dancers from GDance~\citep{le2023music} ($2{,}877$ clips, $16.2$\,h). 
Multi-person clips make up $42.7\%$ of the clips and $35.2\%$ of the hours in total.  
Every clip is accompanied by per-subject instance maps, and every multi-person clip by rendered identity-reference images.

\paragraph{Motion preprocessing.}
All source motions are converted into SMPL-X parameter sequences and standardized to 30 fps.
We remove unsuitable motions including static segments, excessive global displacement, and rotation artifacts.
For each clip, we retain sub-windows whose horizontal root trajectory remains within 8\,m.
For multi-person sequences, we preserve interaction geometry by jointly centering subjects and reject clips with severe inter-person mesh penetration.

\paragraph{Rendering pipeline.}
All videos are rendered with Unreal Engine~5.3~\citep{epicgames_unrealengine5} using the BEDLAM rendering framework~\citep{black2023bedlam, tesch2026bedlam2}.
Filtered motion clips are converted from SMPL-X parameters into per-frame vertex animations and rendered with diverse avatar and scene configurations.
For single-person clips, cameras use near-frontal views and smoothly track the subject motion.
For multi-person clips, a static camera captures the entire group.
The two appearance variants of each pair share identical motion, camera trajectory, and temporal length while differing in subject identity, clothing, hairstyle, body shape, and scene appearance.
Instance masks are generated by rasterizing the corresponding posed SMPL-X geometry with identical camera parameters.
For multi-person clips, identity references are rendered separately in canonical standing poses while preserving the target identities.

\paragraph{Assets.}
Human appearances are constructed from BEDLAM~\citep{black2023bedlam} and BEDLAM~2.0~\citep{tesch2026bedlam2}, including diverse body shapes, garments, hairstyles, and skin appearances.
Each pair is rendered with two distinct identity configurations without sharing identity-specific assets.
Body-shape variation is introduced by perturbing SMPL-X shape coefficients, producing differences in height and body proportions between paired subjects.
Scene environments are sampled from $357$ HDRI maps from Poly Haven~\citep{polyhaven}, covering diverse indoor and outdoor environments.

\clearpage
  
\section{Additional Details on Training WL2P}
\label{app:training_details}

We update only the self-attention modules of the Wan2.1-I2V-14B backbone in all $40$ transformer blocks, including the query, key, value, and output projections.
This amounts to $4.20$B trainable parameters out of $16.39$B ($25.6\%$). 
For optimization, we train with AdamW (weight decay $0.01$) at a constant learning rate of $1\times10^{-5}$ without warmup, gradient clipping at $1.0$, one sample per GPU on four NVIDIA B200 GPUs (effective batch size $4$, no gradient accumulation), bf16 mixed precision, and DeepSpeed ZeRO stage~2.
Following the official Wan-Animate-14B configuration, we use a single fixed text prompt for all training samples and at inference. 
Classifier-free guidance is applied only to the text condition with a scale of $5.0$, while the reference image and driving video conditions are shared between the conditional and unconditional branches.

\paragraph{IBS details.}
Subject labels are assigned on the token grids of the reference image and of every target latent frame ($16$ pixels per token): a token receives the label of the subject covering the largest part of its $16\times16$ region if that fraction is at least $\gamma=0.7$, and is left unlabeled
otherwise (background, boundaries, occluded areas). 
IBS is active only for samples in which at least two subjects survive labeling in \emph{both} the reference and the target.
Queries and keys are taken after RoPE from the self-attention modules, and the loss is computed per attention head and averaged over heads and blocks as in Eq.~\ref{eq:ibs_loss}.
No extra parameters are introduced and inference is unchanged. 
For efficiency, at most $256$ labeled target tokens per subject are sampled as queries in each sample. 
The keys are all labeled reference-image tokens, and the softmax in Eq.~\ref{eq:ibs_attention} is taken over these keys only. 
The loss is weighted by $\beta=0.05$ and applied only at flow-matching timesteps $\tau \ge 0.6$ in Eq.~\ref{eq:total_loss}.

\section{Additional Details on MotionTwin-Bench}
\label{app:benchmark}

\paragraph{Benchmark construction.}
MotionTwin-Bench is constructed from held-out clips that are strictly separated from the training set.
The benchmark follows the same rendering pipeline and appearance generation process as the training set.
For the multi-human motion scenarios, we select challenging interaction windows with large motion, frequent turning, and close inter-person interactions, while removing sequences with severe mesh interpenetration.
Per-subject instance maps are generated for both variants to provide ground-truth identity correspondence.

\paragraph{Reference images.}
For each multi-person benchmark pair, the identity-reference image is rendered using the same camera, HDRI, and lighting configuration as the target video to maintain appearance consistency.
Subjects are placed according to their initial ordering in the target window and rendered in canonical standing poses facing the camera.
The reference image and its instance map are generated using the same geometry-based procedure as the target videos, providing ground-truth identity correspondence.

\paragraph{Identity assignment metrics.}
For multi-person windows we evaluate whether each generated subject carries the intended reference
identity. 
Identity Assignment Accuracy (IAA) is the fraction of frames in which \emph{all} subjects are assigned to their correct reference identity. $\mathrm{IAA}_{\mathrm{cross}}$ restricts the average to frames in which at least two subject bounding boxes overlap (IoU $>0.1$), i.e., frames
with inter-person proximity or occlusion, where identity-motion binding is most challenging. 

\paragraph{Assignment procedure.}
Since the generated motion follows the driving video, the ground-truth per-subject instance maps of the target window locate each subject in every generated frame; the question is only \emph{which} reference identity appears there. 
For each reference subject, we compute an appearance signature from the reference image restricted to that subject's instance region, and for each generated subject region the same signature from the generated frame. 
The signature is $27$-dimensional: the mean CIELAB color ($3$) and an $8$-bin histogram of each RGB channel ($24$), both normalized. 
We then find the one-to-one assignment between generated regions and reference subjects that minimizes the total $L_2$ signature distance (exhaustive over permutations, $P\le4$), and a frame counts as correct only if this assignment is the identity mapping. 
Because the signatures are computed inside ground-truth regions, the procedure is insensitive to pose and to small misalignments and measures only whether the right appearance was placed on the right trajectory.

\begin{table}[t]
    \centering
    \vspace{-4mm}
    \caption{
        Quantitative evaluation compared to state-of-the-art methods on single-person animation. MotionTwin-Bench metrics are computed on the 75 single-subject windows, and VBench metrics on the 36 single-subject real-world clips.
    }
    \label{tab:single_person}
    \setlength{\tabcolsep}{4.0pt}
    \renewcommand{\arraystretch}{1.15}
    \resizebox{\linewidth}{!}{%
    \begin{tabular}{l|cccc|cc|c}
        \toprule
        \multirow{2}{*}{\vspace{-5pt}Method}
        & \multicolumn{4}{c|}{MotionTwin-Bench}
        & \multicolumn{2}{c|}{Vbench}
        & Vbench++ \\
        \cmidrule{2-8}
        & \makecell{PSNR$\uparrow$}
        & \makecell{SSIM$\uparrow$}
        & \makecell{LPIPS$\downarrow$}
        & \makecell{FVD$\downarrow$}
        & \makecell{V-Quality$\uparrow$}
        & \makecell{F-Quality$\uparrow$}
        & \makecell{I2V-Quality$\uparrow$}
        \\
        \midrule
        MultiAnimate~\citep{Hu_2026_CVPR}
        & 20.126
        & 0.5377
        & \underline{0.2283}
        & 317.51
        & 80.32
        & 81.64
        & 85.61 \\

        Wan-Animate 2~\citep{wang2026wan}~~
        & \underline{20.926}
        & \underline{0.5846}
        & 0.2447
        & \underline{216.85}
        & 81.84
        & 83.11
        & 86.91 \\

        SCAIL~\citep{yan2026scail}
        & 18.389
        & 0.4571
        & 0.2654
        & 263.70
        & 80.97
        & 82.31
        & 86.33 \\

        SCAIL-2~\citep{yan2026scail2}
        & 18.754
        & 0.4608
        & 0.2877
        & 273.18
        & \underline{82.49}
        & 84.10
        & 88.91 \\

        \midrule
        Wan-Animate-14B~\citep{cheng2025wan}
        & 17.691
        & 0.4621
        & 0.3353
        & 267.81
        & 81.25
        & 82.53
        & 86.39 \\

        SteadyDancer~\citep{zhang2025steadydancer}
        & 19.738
        & 0.5035
        & 0.2830
        & 233.54
        & 80.86
        & \underline{84.31}
        & \textbf{94.67} \\

        \midrule
        \textbf{WeLike2Party~(Ours)}
        & \textbf{22.328}
        & \textbf{0.6316}
        & \textbf{0.1903}
        & \textbf{147.70}
        & \textbf{83.41}
        & \textbf{85.09}
        & \underline{90.15} \\

        \bottomrule
    \end{tabular}%
    }
    \vspace{-3mm}
\end{table}

\label{tab:single_person_appendix}

\section{Additional Experiments}
\label{sec:additional_experiments}

\noindent\textbf{Single-person evaluation.} We evaluate WL2P on the single-person subsets of MotionTwin-Bench and the real-world benchmark to verify that its multi-person design does not compromise single-person animation.
We additionally compare with Wan-Animate-14B~\citep{cheng2025wan} and SteadyDancer~\citep{zhang2025steadydancer}, which are specifically designed for single-person animation.
Since subject fidelity and identity binding metrics are formulated for multi-person scenarios, we report only full-frame fidelity metrics. 

As shown in Tab.~\ref{tab:single_person_appendix}, WL2P achieves the best reconstruction fidelity and FVD on the single-person subset of MotionTwin-Bench, including comparisons with single-person-only methods.
On real-world clips, WL2P achieves the best video and frame quality, while remaining competitive in I2V quality.
These results indicate that WL2P retains strong single-person animation performance despite being designed for multi-person settings.

\begin{figure}[t]
    \centering
    \includegraphics[width=\linewidth]{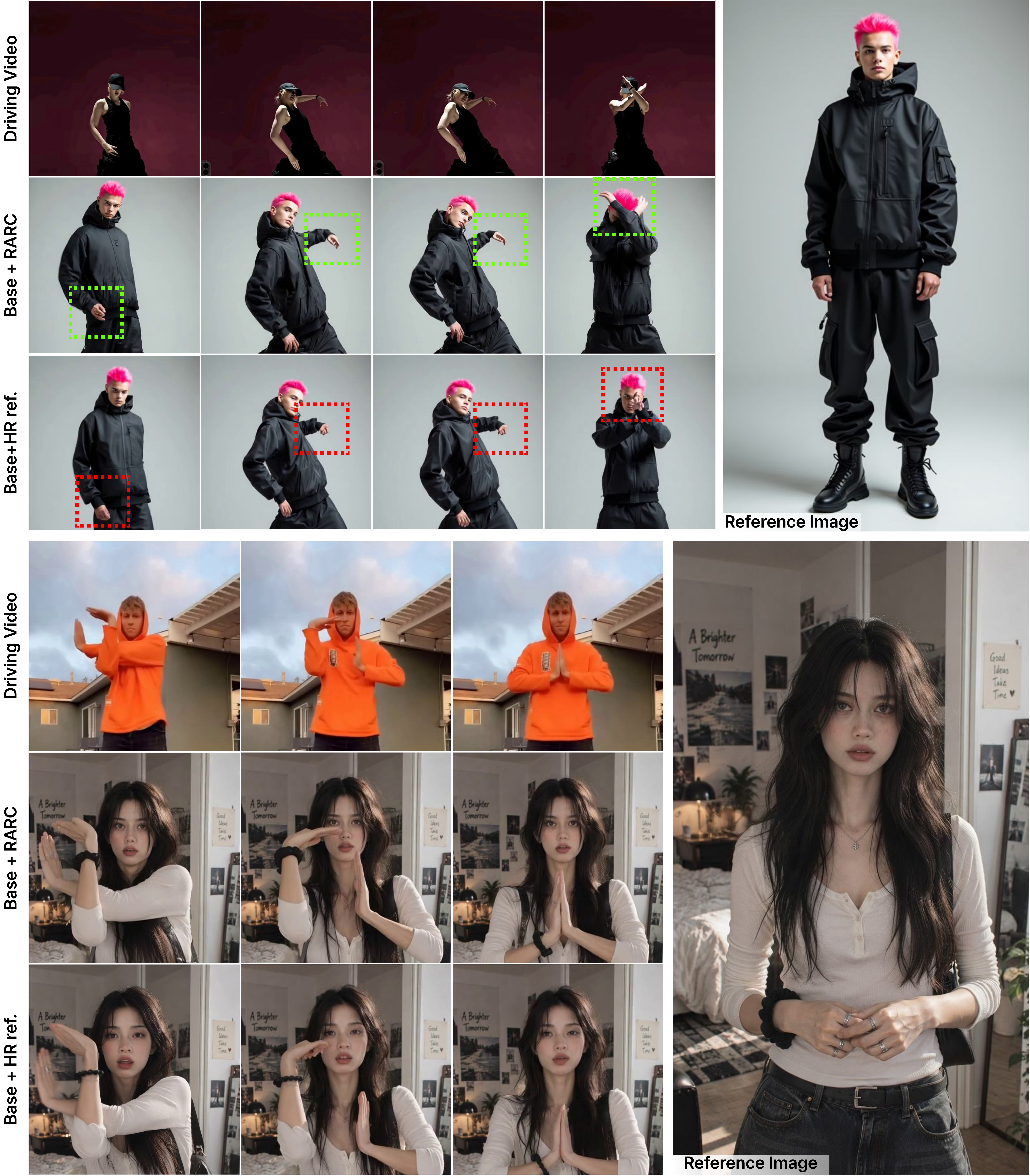}
    \vspace{-7mm}
    \caption{\textbf{Qualitative ablation of RARC.} Both variants receive the same high-resolution reference image and driving video and differ only in how reference tokens are positioned. With RARC, fingers are rendered distinctly and the face stays intact under fast motion (top, green), whereas assigning native integer RoPE coordinates (Base+HR ref.) yields merged fingers (top, red). The second example (bottom) shows the same trend for hand gestures near the face, where Base+HR ref. blurs the fingers and exhibits identity drift while RARC preserves both.}
    \label{fig:abl_rarc}
\end{figure}

\noindent\textbf{Ablation study on RARC.} Fig.~\ref{fig:abl_rarc} compares RARC with the variant that encodes the reference at the same high resolution but keeps native integer RoPE coordinates.
Simply enlarging the reference does not recover fine details.
The extra tokens fall outside the spatial range seen during pretraining, so the model cannot reliably localize which reference region each target token should attend to.
As a result, fingers are merged into a blob (red boxes), even though the overall body motion is transferred.
With RARC, the same tokens are mapped into the target coordinate range, and the model renders distinct, correctly separated fingers and preserves face appearance throughout the gesture (green boxes).
The second example shows the same behavior as the first one. 
This confirms that the gain comes from placing the denser reference grid within the pretrained RoPE range rather than from higher reference resolution alone.

\clearpage

\section{Additional Qualitative Results}

\begin{figure}[h]
  \centering
  \includegraphics[width=\linewidth]{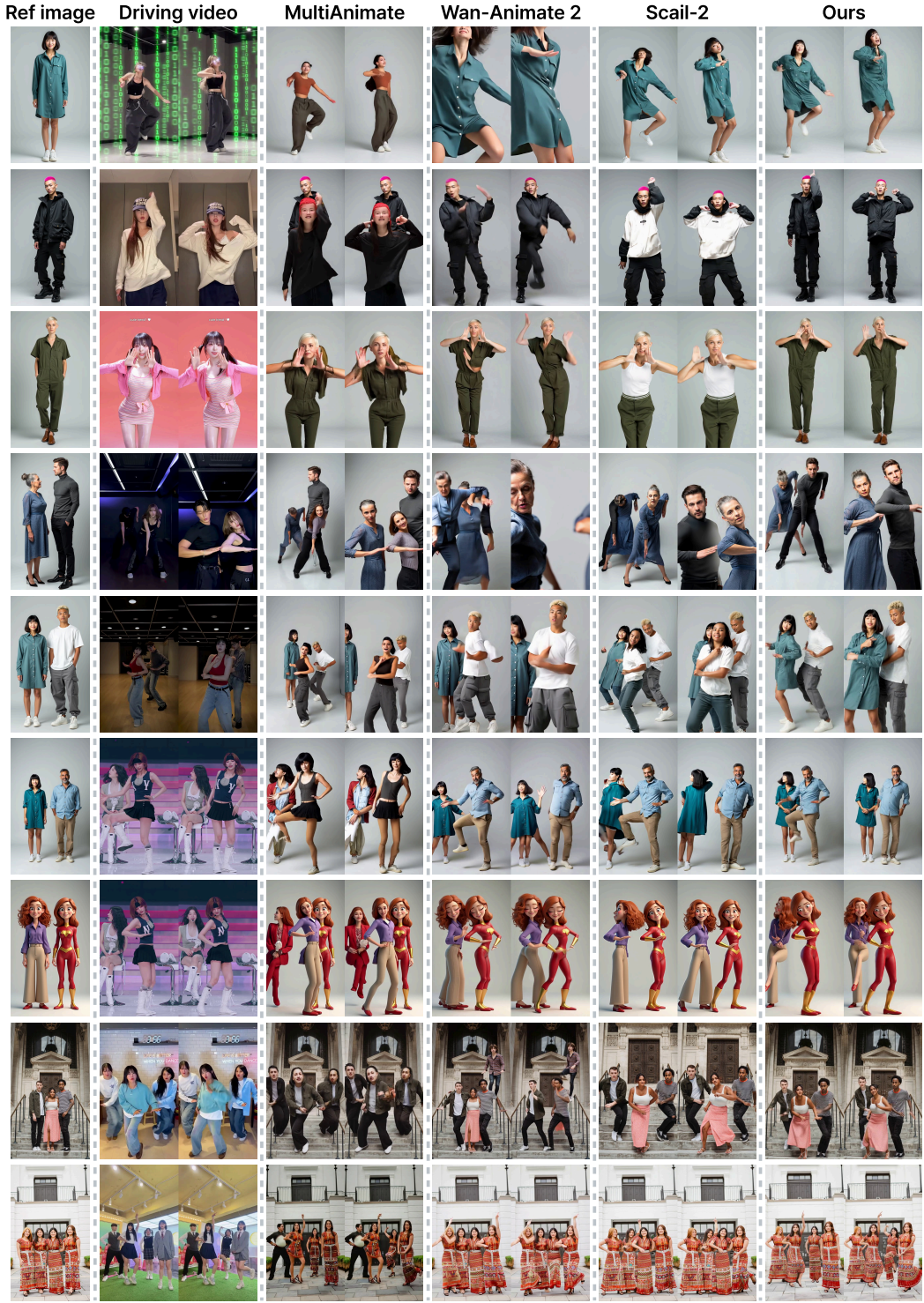}
  \caption{\textbf{Additional qualitative results.}}
  \label{fig:addquali1}
\end{figure}

\begin{figure}[t]
  \centering
  \includegraphics[width=\linewidth]{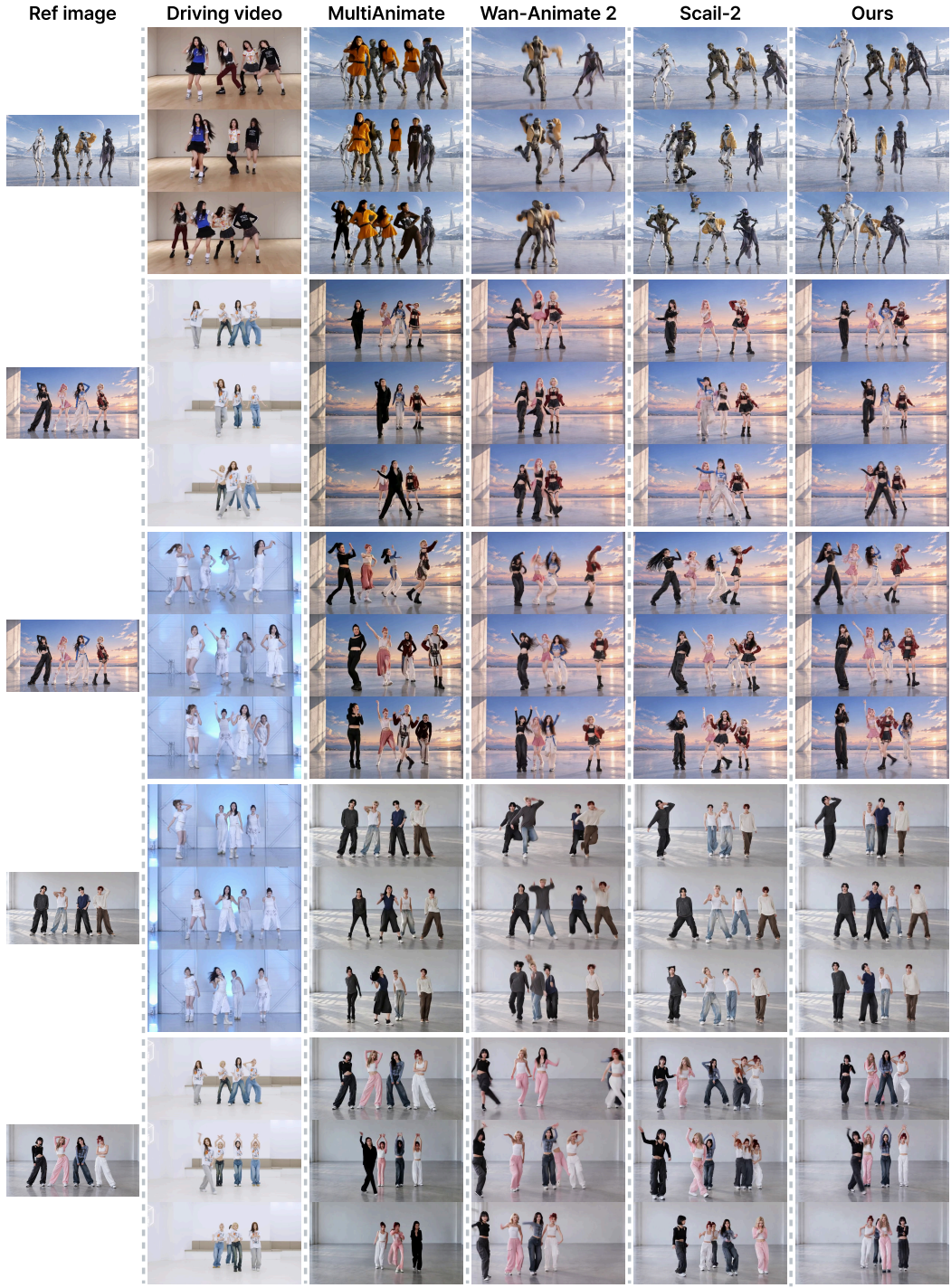}
  \caption{\textbf{Additional qualitative results.}}
  \label{fig:addquali2}
\end{figure}

\begin{figure}[t]
  \centering
  \includegraphics[width=\linewidth]{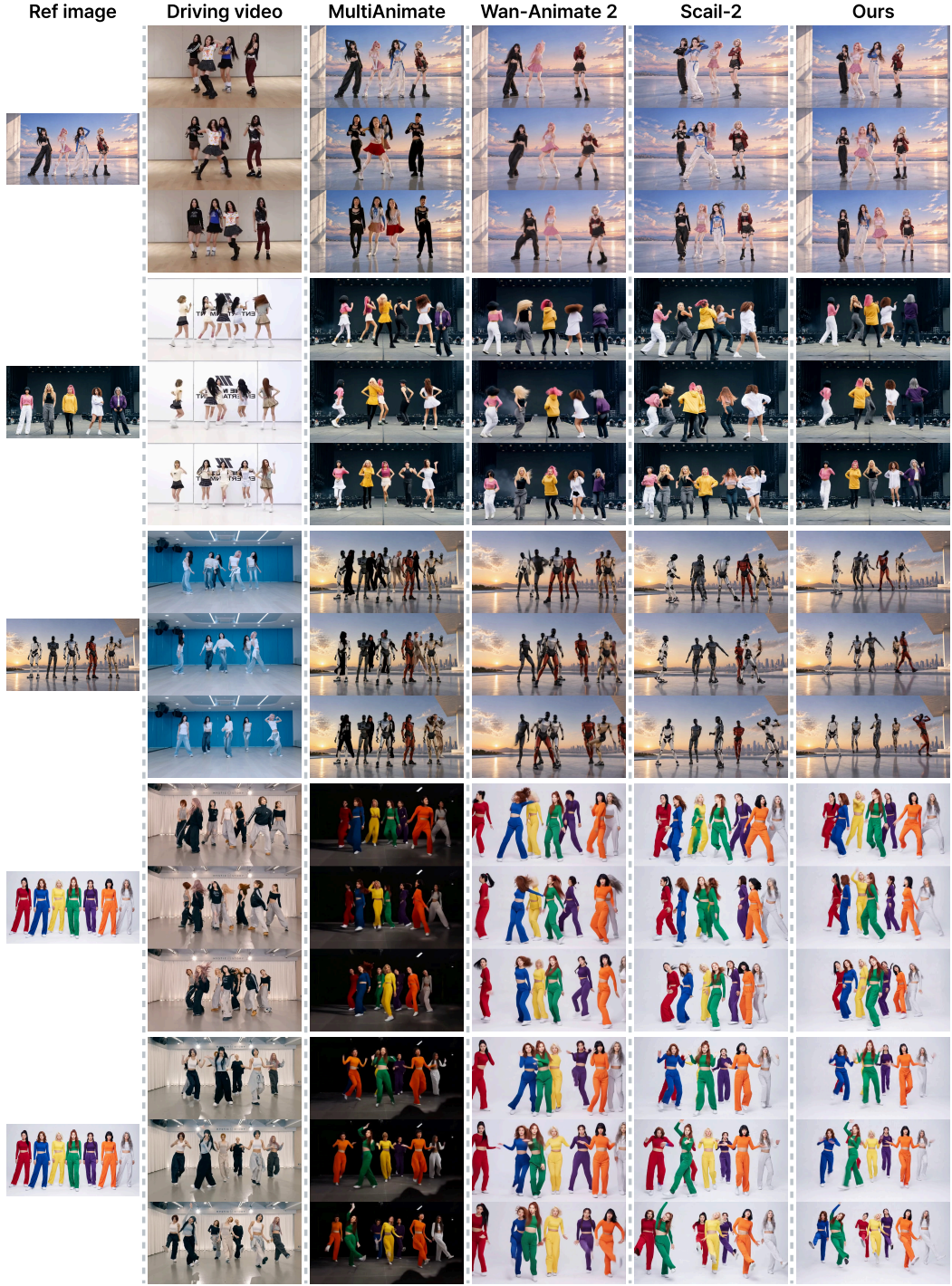}
  \caption{\textbf{Additional qualitative results.}}
  \label{fig:addquali3}
\end{figure}

\end{document}